\documentclass[lettersize,journal]{IEEEtran}
\usepackage{amsmath,amsfonts}
\usepackage{algorithmic}
\usepackage{algorithm}
\usepackage{array}
\usepackage[caption=false,font=normalsize,labelfont=sf,textfont=sf]{subfig}
\usepackage{textcomp}
\usepackage{stfloats}
\usepackage{url}
\usepackage{verbatim}
\usepackage{graphicx}
\usepackage{cite}
\usepackage{multirow}

\def\eg{\emph{e.g.}} 

\def\ie{\emph{i.e.}}

\def\etc{\emph{etc.}}

\def\etal{\emph{et al.}}

\usepackage{color}
\usepackage[pagebackref=true,breaklinks=true,letterpaper=true,colorlinks,bookmarks=false]{hyperref}

\begin{document}

\title{VDTR: Video Deblurring with Transformer}

\author{
    Mingden~Cao,
    Yanbo~Fan,
    Yong~Zhang,
    Jue~Wang,
    Yujiu~Yang
\thanks{M. Cao and Y. Yang are with Tsinghua Shenzhen International Graduate School, Tsinghua University, Shenzhen 518055, China. Y. Fan, Y. Zhang and J. Wang are with Tencent AI Lab, Shenzhen 518054, China. 

Y. Fan and Y. Yang are corresponding authors~(Emails: yanbofan@tencent.com, yang.yujiu@sz.tsinghua.edu.cn).

}
}

\markboth{ }%
{Shell \MakeLowercase{\textit{et al.}}: A Sample Article Using IEEEtran.cls for IEEE Journals}

\maketitle

\begin{abstract}
Video deblurring is still an unsolved problem due to the challenging spatio-temporal modeling process. While existing convolutional neural network-based methods show a limited capacity for effective spatial and temporal modeling for video deblurring. This paper presents VDTR, an effective Transformer-based model that makes the first attempt to adapt Transformer for video deblurring. VDTR exploits the superior long-range and relation modeling capabilities of Transformer for both spatial and temporal modeling. However, it is challenging to design an appropriate Transformer-based model for video deblurring due to the complicated non-uniform blurs, misalignment across multiple frames and the high computational costs for high-resolution spatial modeling. 
To address these problems, VDTR advocates performing attention within non-overlapping windows and exploiting the hierarchical structure for long-range dependencies modeling. For frame-level spatial modeling, we propose an encoder-decoder Transformer that utilizes multi-scale features for deblurring. For multi-frame temporal modeling, we adapt Transformer to fuse multiple spatial features efficiently. Compared with CNN-based methods, the proposed method achieves highly competitive results on both synthetic and real-world video deblurring benchmarks, including DVD, GOPRO, REDS and BSD. We hope such a Transformer-based architecture can serve as a powerful alternative baseline for video deblurring and other video restoration tasks. The source code will be available at \url{https://github.com/ljzycmd/VDTR}.
\end{abstract}


\section{Introduction}
\IEEEPARstart{U}{ndesired} blurs are often unavoidable because of fast-moving objects or the shaking camera when trying to acquire a video with hand-held devices. The blurs significantly deteriorate the visual quality and the video information. Video deblurring that tries to restore the latent sharp frames from a blurry video has many practical applications~\cite{dai2006tracking, kupyn2018deblurgan, lin2020learning}. It remains a challenge because the blurs vary both spatially and temporally. A large receptive field is required to handle the non-uniform blurs within each frame. Meanwhile, modeling the temporal variations among consecutive frames is essential for video deblurring, which can utilize the complementary information from adjacent frames for better reconstruction. 

\begin{figure}[t]
    \centering
    \includegraphics[width=\linewidth]{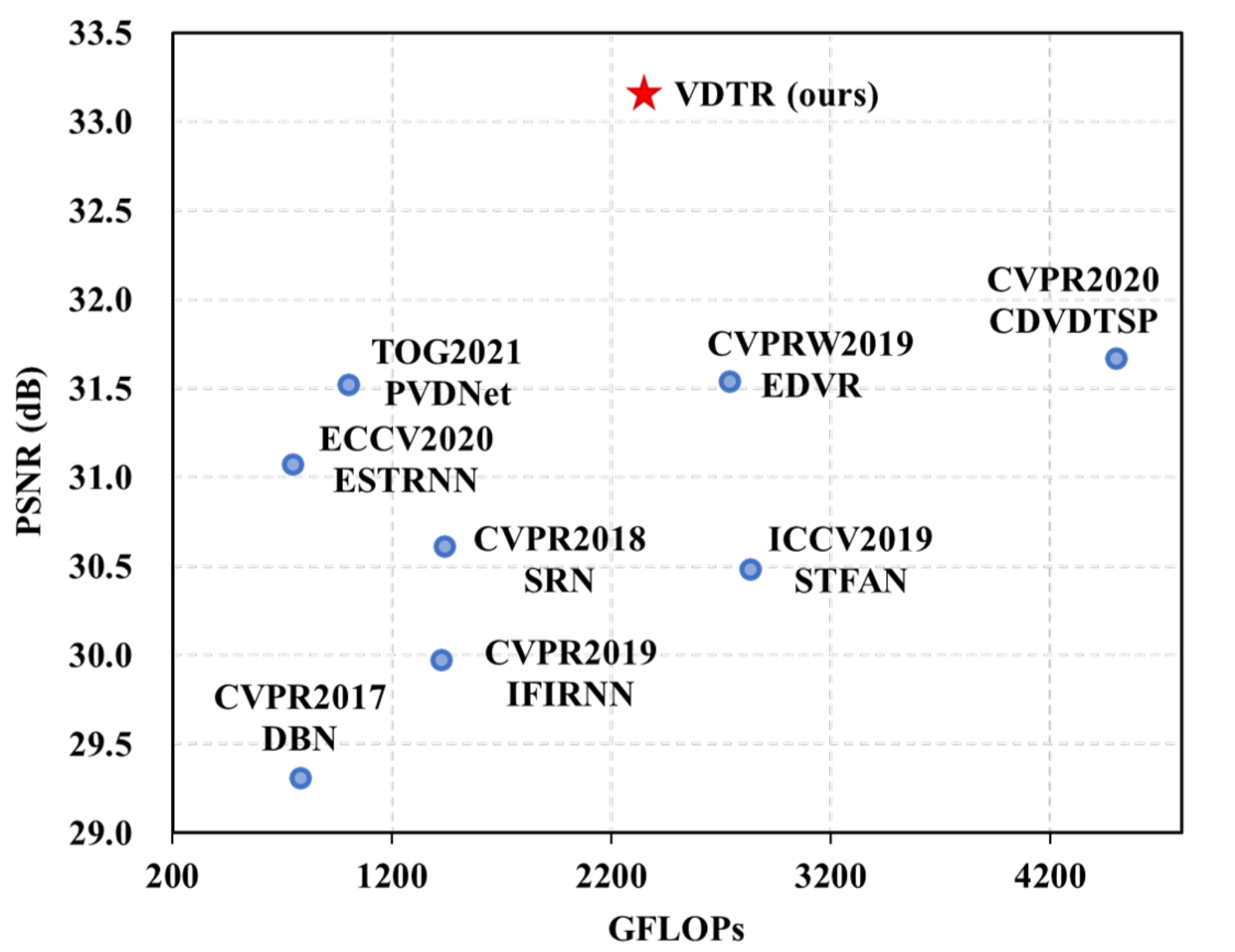}
    \caption{Comparison between the proposed Transformer-based VDTR and existing CNN-based approaches in terms of the PSNR ($\uparrow$) and GFLOPs ($\downarrow$). These data are tested on the popular video deblurring dataset GOPRO~\cite{nah2017deep}. The GFLOPs are calculated under the resolution $720\times1280$ of the input frame. We see that VDTR significantly outperforms the counterparts. }
    \label{fig:comparison_teaser}
\end{figure}

Most existing methods adopt specific architectural designs for long-range spatial modeling to model the frame-level spatial relationships for deblurring. For example, the works of~\cite{nah2017deep, tao2018scale, zhang2019dhmpn} exploited the multi-stage architecture to get a large receptive field relatively with a coarse-to-fine strategy. Yet these methods often hold a long processing time because they gradually restore the latent image from the smaller scale images. 
With the help of dilated convolution~\cite{yu15dilated} and deformable convolution~\cite{dai2017deformable}, the works of~\cite{yuan2020efficient, purohit2020region} proposed to deal with the blurs caused by large motion by modeling the long-range dependencies more efficiently. Nevertheless, these models often suffer from an unstable training process, which greatly deteriorates deblurring performance. For the temporal modeling, which tries to utilize the information from neighboring frames, the difficulty is to build effective correspondence among the misaligned frames or features. Many methods try to ease this problem by first aligning these frames first with optical flow~\cite{su2017deep, pan2020cascaded}, deformable convolution~\cite{wang2019edvr}, or dynamic convolution~\cite{zhou2019spatio}, then aggregating the aligned features for a better reconstruction. However, it is still challenging to align multiple frames, especially when incredibly blurry.

Recently, Transformer~\cite{vaswani2017attention} has shown powerful long-range and relation modeling capabilities in natural language processing~(NLP) and is gradually introduced into computer vision~(CV), showing impressive performance in object detection~\cite{carion2020detr, zhu2021deformdetr} and image classification~\cite{dosovitskiy2020vit, touvron2020deit}, served as Vision Transformer. The superior long-range and relation modeling capacities of Transformer inspire us to explore it for video deblurring. However, directly applying the architecture of Vision Transformer designed for detection or classification may not be effective for video deblurring for the following reasons. (1) The large patch partition is inappropriate for deblurring, which requires finer-grained details for high-quality image reconstruction; (2) The non-uniform blurs make it hard to handle by a single-scale representation; (3) The misalignment across multiple frames is challenging for temporal modeling; (4) The computational complexity of Vision Transformer grows quadratically corresponding to the frame size. The high costs prevent self-attention operations from the application in high-resolution scenarios.

To move beyond these limitations, in this paper, we propose a novel Video Deblurring TRansformer (VDTR) that takes advantage of the long-range and relation modeling characteristics of Transformer for video deblurring. Firstly, to maintain the representations in high resolution for high-quality reconstruction, VDTR employs $4\times4$ patch size for patch embedding to generate fine-grained image embedding. Then we propose a Transformer-based encoder-decoder network for frame-level spatial modeling, extracting multi-scale features to handle the blurs caused by different degrees of movement. Secondly, as for the misalignment in multiple frames, we design a temporal Transformer to build the correspondence across frames within a local patch, showing a better performance on aggregating complementary information. Lastly, we advocate a local Transformer, which integrates local window-based attention rather than the global one for high-resolution processing, which reduces the computational complexity of the Transformer from quadratic to linear, corresponding to the frame size. With the help of shifted window attention~\cite{liu2021swin}, the local Transformer can maintain the long-range dependency modeling capacity. 

We conduct numerous experiments on the popular synthetic and real-world video deblurring benchmarks, including DVD~\cite{su2017deep}, GOPRO~\cite{nah2017deep}, REDS~\cite{Nah_2019_CVPR_Workshops_REDS} and BSD~\cite{zhong2020estrnn}. The comparisons shown in Fig.~\ref{fig:comparison_teaser} reveal that VDTR achieves much more competitive video deblurring performance in terms of PSNR than state-of-the-art CNN-based methods. We hope such a Transformer-based model can serve as an alternative baseline model for efficient video deblurring by utilizing the superior modeling capacities of Transformer. 

Our contributions can be summarized as follows:
\begin{itemize}
\item We propose to utilize the superior long-range and relation modeling capacities of Transformer for video deblurring.
\item We design a Transformer-based encoder-decoder network to efficiently extract the multi-scale frame-level spatial features to tackle non-uniform blurs.
\item We also propose a novel temporal Transformer to aggregate sharp information from misaligned frames. 
\item Extensive experimental results on both synthetic and real-world video deblurring benchmarks demonstrate the effectiveness of VDTR and can be an alternative to CNNs. 
\end{itemize}

\begin{figure*}[t]
\centering
\includegraphics[width=\linewidth]{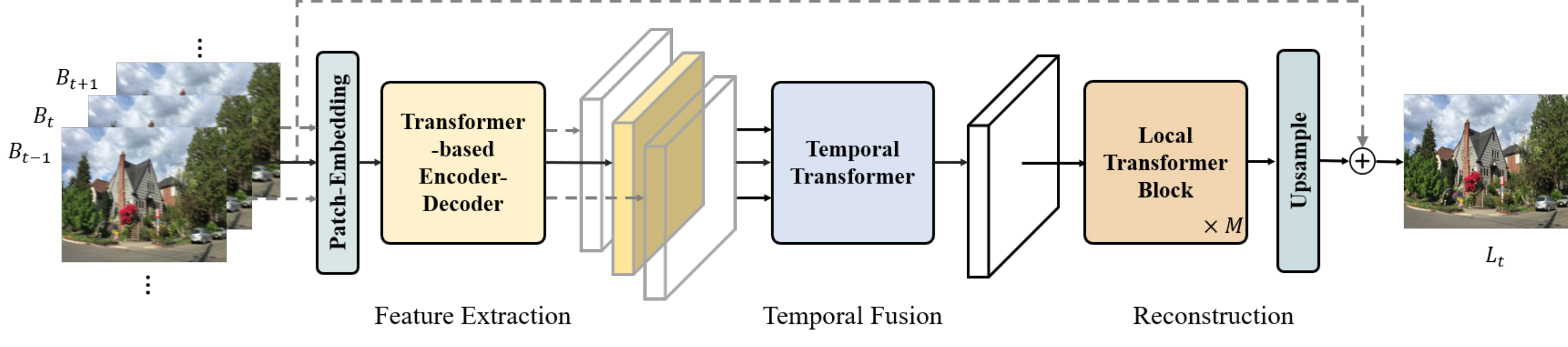}
\caption{The main architecture of proposed VDTR. $2N+1$ consecutive blurry frames are fed into VDTR and output one sharp frame corresponding to the central blurry frame. The proposed Transformer-based encoder-decoder network first extracts the hierarchical features of each frame. Then multiple frame-level features are fused by the temporal Transformer, which aggregates sharp information from neighboring frames. Last, the temporal-fused feature is reconstructed into a high-quality latent frame with local Transformer blocks. VDTR fully exploits pure Transformer for video deblurring and achieves favorable performance against state-of-the-art CNN-based methods.}
\label{fig:model_arch}
\end{figure*}

\section{Related Work}
\subsection{Video Deblurring}
Video deblurring requires effective spatio-temporal modeling to extract complementary information existing in neighboring frames to help better restoration. Early video or multi-frame deblurring approaches are often based on the observation that different frames are not equally blurred in a video. So methods directly find and select sharp pixels from multiple frames and serve as the final output~\cite{joshi2010seeing, law2006lucky}. In~\cite{matsushita2006full, cho2012video}, researchers first align multiple frames with homography and path-based alignment, and then aggregate sharp pixels to generate a latent sharp frame. Some methods also estimate spatio-temporally varying blur kernels with motion modeling, and then the kernels are used for deconvolution~\cite{hyun2015generalized}. However, these methods often require much computational costs and human priors to optimize the energy function for convergence, greatly limiting its practical applications.

Compared to traditional methods, deep learning-based methods have shown more competition in both restored video quality and inference time. Inspired by the success of Encoder-decoder architecture in image deblurring~\cite{su2017deep, tao2018scale, purohit2020region}, some video deblurring methods adopt such architecture for feature extraction~\cite{wang2019edvr}.  Su~\etal~\cite{su2017deep} proposed an encoder-decoder-based video deblurring network that stacks multiple consecutive frames as input, directly outputing the restored latent image. Recurrent neural networks~(RNNs) are adopted to exploit the temporal information and improve the video deblurring performance with efficient spatial-temporal learning~\cite{kim2017online, nah2019recurrent, zhou2019spatio, zhong2020estrnn}. The extracted features or restored frames are usually used to enhance the current frame with complementary temporal information in these models. Zhang~\etal~\cite{zhang2018adversarial} also employed 3D CNN to capture spatio-temporal information to help the central blurry frame restoration with adversarial training. 

To improve the deblurring performance further, researchers proposed some extra multiple frames aligning methods for more effective temporal modeling. Kim~\etal~\cite{kim2018sttn} proposed a spatio-temporal flow to establish correspondence across several frames for video restoration, which transfers the spatio-temporal information from multiple frames to the current frame. The optical flow between reference and neighboring frames is also used to estimate the motion information for restoring the latent frames with a temporal sharpness prior~\cite{pan2020cascaded, xiang2020deep}. Deformable and dynamic convolutions are also applied to align the adjacent frame features in~\cite{wang2019edvr} and~\cite{zhou2019spatio} implicitly, which achieved better deblurring performance. These methods continuously improved deblurring performance. However, the aligning performed among multiple frames is not accurate when the video is extremely blurry. Our method adopts Transformer to directly build the correspondence among consecutive temporal features, which can model the spatio-temporal information for deblurring adaptively.

\subsection{Vision Transformer}
Transformers are transferred from NLP to CV in many vision tasks by integrating the attention mechanism or the full Transformer as a powerful module and have achieved much success~\cite{wang2018non, ramachandran2019stand, carion2020detr}. For example, DETR~\cite{carion2020detr} adopted a full Transformer architecture for relationship building and refining, which can directly output the detection results without any post-process. The pure Transformer-based architecture is also designed for classification~\cite{dosovitskiy2020vit, touvron2020deit, han2021tit, touvron2021cait, chen2021crossvit, shu2021adder}, achieving highly competitive accuracy compared to CNNs. A recent breakthrough in Vision Transformer is the hierarchical design in Pyramid Vision Transformer (PVT)~\cite{wang2021pvt} and Swin Transformer~\cite{liu2021swin}. They developed general pyramid-like architecture backbones for various downstream tasks, \eg, detection and segmentation \etal This architecture can extract hierarchical features of the input image, like the CNN-based backbones done before. 

These Transformer-based backbones surpass the existing CNN-based backbone, inspiring researchers to explore the application of Transformer in the restoration field. Chen~\etal utilized Vision Transformer for image restoration~\cite{chen2020ipt}, which adopts multi-heads and multi-tails to adapt for different image restoration tasks. However, it needs a large amount of data for pretraining. Further, SwinIR~\cite{liang2021swinir} modified Swin Transformer block for image super-resolution, denoise, \etc, achieving a new state-of-the-art with much fewer parameters. However, there are only few explorations for video restoration tasks.
In this paper, we propose VDTR, making the first attempt to adopt Transformer to video deblurring to the best of our knowledge.

\section{Video Deblurring with Transformer}
In this section, we elaborate on the exploration of the Transformer introduced to the video deblurring task. Specifically, we first introduce the overall architecture of VDTR, then present its critical designs in detail. 

\subsection{Overall Architecture}
We aim to utilize the long-range and relation modeling of Transformer to effectively model the spatially and temporally varying blurs in the blurry videos.
The overall architecture of the proposed VDTR model is presented in Fig.~\ref{fig:model_arch}, which mainly consists of three parts: (1) frame-level spatial feature extraction; (2) temporal modeling across multiple frames; (3) latent frame reconstruction. More details are introduced as follows.

To better exploit the sharp information existing in the neighboring frames, VDTR takes $2N+1$~($N$ is the number of future and past frames at current time $t$) consecutive blurry frames $\{B_{t+i}\}_{i=-N}^N$ as input, and output the central latent frame $L_t$ corresponding to the reference frame $B_t$ with an end-to-end manner. Following the image-to-patch method in ViT~\cite{dosovitskiy2020vit}, each blurry frame $B_{t+i} \in \mathbb{R}^{3\times H \times W}$~($H$ and $W$ are the height and width of input frames) is decomposed into non-overlapping patches first, then they are flatten and linear projected to embedding vectors with dimension $d$~($d=256$ in VDTR) by a Patch-Embedding operation. Thus the blurry frame embedding $B^{e}_{t+i}$ are with the shape of $\mathbb{R}^{d \times \frac{H}{p} \times \frac{W}{p}}$, where $p$ is the patch size during patch embedding. We employ $4 \times 4$ patch partition to generate finer-grained representations for high-quality restoration. 

After obtaining the initial frame-level embedding $B^{e}_{t+i}$, the frame-level features are further extracted by the well-designed feature extractor $G_f$. To better perform effective frame-level spatial modeling for non-uniform motion blurs removal, the extractor with large-range and multi-scale representative capacity is required. To achieve so, we design a Transformer-based encoder-decoder network to extract multi-scale frame-level spatial representations as:

\vspace{-0.5cm}
\begin{align}
    F_{t+i} = G_{f}(B^e_{t+i}),
\end{align}
where $F_{t+i}$ is the extracted feature corresponding to the $i$-th neighboring blurry frame. $G_f$ learns multi-scale representations to adapt the non-uniform blurs in the video. In addition, to reduce the computational costs, we exploit the local Transformer blocks in $G_f$, which will be introduced later. All input blurry frames share the same feature extractor. 

Next, the extracted frame-level spatial features $\{F_{t+i}\}^{N}_{i=-N}$ are fed into a well-designed temporal Transformer $G_t$ to extract complementary information existing in neighboring frames as:

\vspace{-0.5cm}
\begin{align}
    F_{fused} = G_{t}(\{F_{t+i}\}_{i=-N}^{N}),
\end{align}
where $F_{fused}$ is the feature aggregated from multiple frames. 

At last, the temporally fused features are reconstructed to the latent sharp frame $L_t$ with a global residual learning strategy:

\vspace{-0.5cm}
\begin{align}
    L_t = G_{r}(F_{fused}) + B_t,
\end{align}
where $G_r$ is the reconstruction process, containing $M$ local Transformer blocks~($M=20$ in our experiment), two pixel-shuffle layers for upsampling, and a linear projection layer to project the upsampled features into a RGB map.

\begin{figure}[!t]
    \centering
    \includegraphics[width=\linewidth]{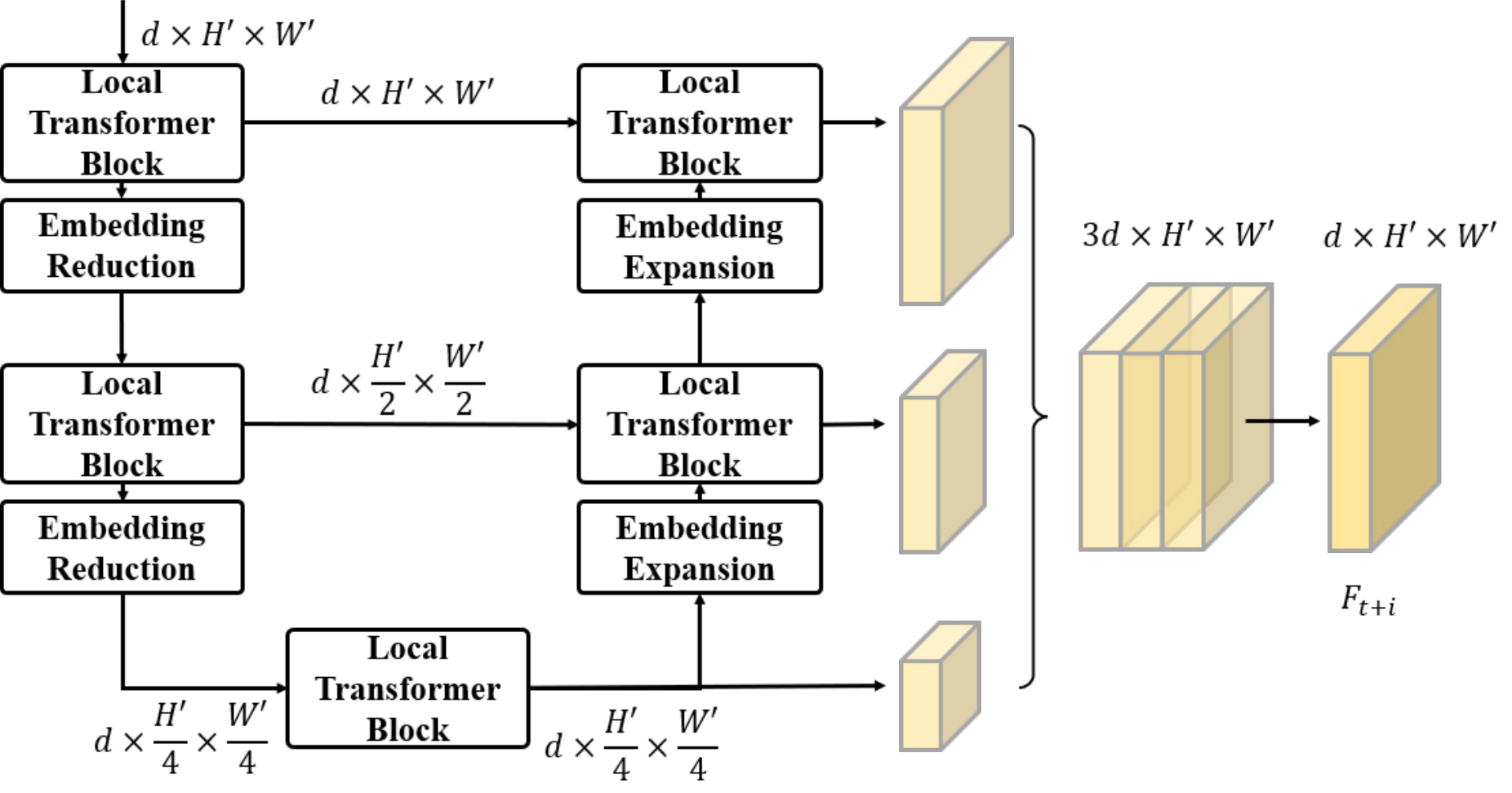}
    \caption{Illustration of Transformer-based encoder-decoder network, which extract multi-scale features to adapt different degrees of blurs. The smaller scale features are first upsampled to the largest one, and concatenate over the channel dimension, then they are linear projected to the frame-level spatial feature $F_{t+i}$. }
    \label{fig:trans_unet}
\end{figure}

\subsection{Frame-level Feature Extraction with Transformer}
\label{subsec:feature_extraction}
\paragraph{Transformer-based Encoder-decoder Network}
We propose a Transformer-based encoder-decoder network with a multi-scale design shown in Fig.~\ref{fig:trans_unet}, which utilizes multi-scale features to handle the non-uniform blurs caused by different degrees of movement. Specifically, this feature extractor contains an encoder and a decoder sub-network with symmetrical architecture. In each stage of the encoder, the input features are first modeled by local Transformer blocks; then, they are spatial downsampled two times by an Embedding-Reduction operator. The embedding-reduction operator reshapes the input features with shape $\mathbb{R}^{d \times H \times W}$ to the shape $\mathbb{R}^{4d \times \frac{H}{2} \times \frac{W}{2}}$ and then linearly projects it to $\mathbb{R}^{d \times \frac{H}{2} \times \frac{W}{2}}$. 
All stages are stacked to extract the hierarchical features. For the decoder branch, each stage consists of an Embedding-Expansion operator followed by the local Transformer blocks. Similarly, the Embedding-Expansion first increases the number of the input feature map channels four times by linear projection and then reshapes it to a spatial upsampled feature map. These features from the previous decoder stage are then fused with the feature maps in the corresponding encoder stage for further feature refinement by the local Transformer blocks. 

The decoder would output multi-scale hierarchical representations. We upsample the smaller ones to the largest spatial size, then concatenate the upsampled features and linearly project them to the output features$F_{t+i}$, which are later used for temporal modeling. The detailed process is shown in Fig.~\ref{fig:trans_unet}. Through this Transformer-based encoder-decoder network, the output features contain multi-scale information, which is beneficial to deal with the blurs caused by different degrees of motion. 

\paragraph{Local Transformer Block}
The original Vision Transformer performs multi-head attention across all spatial positions, owning a quadratic computational complexity corresponding to the image size: $ \mathcal{O} = 2(HW)^2d$, making it unacceptable to handle high-resolution frames, especially video deblurring tasks. To design the Transformer blocks for video deblurring, we advocate performing local attention in non-overlapping windows. 
Let the feature $F \in \mathbb{R}^{d \times H \times W}$ be the input of a certain local Transformer block, it is first partitioned into the non-overlapping windows with spatial size $m \times m$ by a reshape operation. After that, a multi-head self-attention is executed within each window. For the features $X \in \mathbb{R}^{m\times m\times d}$ in a local window, a 2D learnable positional encoding is added to it firstly, then generating the Query, Key and Value by linear projection with transformation matrices $W_q$, $W_k$, and $W_v$:

\vspace{-0.5cm}
\begin{align}
    Q = Z W_q, \quad K = Z W_k, \quad V = Z W_v. 
\end{align}
where $Z = \text{Flatten}(X + LPE)$, $LPE$ denotes 2D learnable positional encoding which concatenates two 1D positional encoding corresponding to the height and width of the window. Then the attention is performed as:

\vspace{-0.5cm}
\begin{align}
    \label{eq:attn}
    \text{Attention}(Q, K, V) = \text{SoftMax}(\frac{QK^T}{\sqrt{d}})V.
\end{align}
The computational complexity of local window-based attention~(W-MSA) is $\mathcal{O}(\text{W-MSA}) = 2m^2HWd$, which grows linearly related to the image size $H\times W$. A local Transformer block further integrates a feed-forward network~(FFN) successively. In this way, the Transformer can maintain the high-resolution representations by vastly reducing computational costs without severely down-sampling the frames or feature maps. 
We also utilize the shift window operation in Swin Transformer~\cite{liu2021swin} to enhance the long-range dependency modeling capacities of Transformer for latent frame reconstruction.

\begin{figure}
    \centering
    \includegraphics[width=\linewidth]{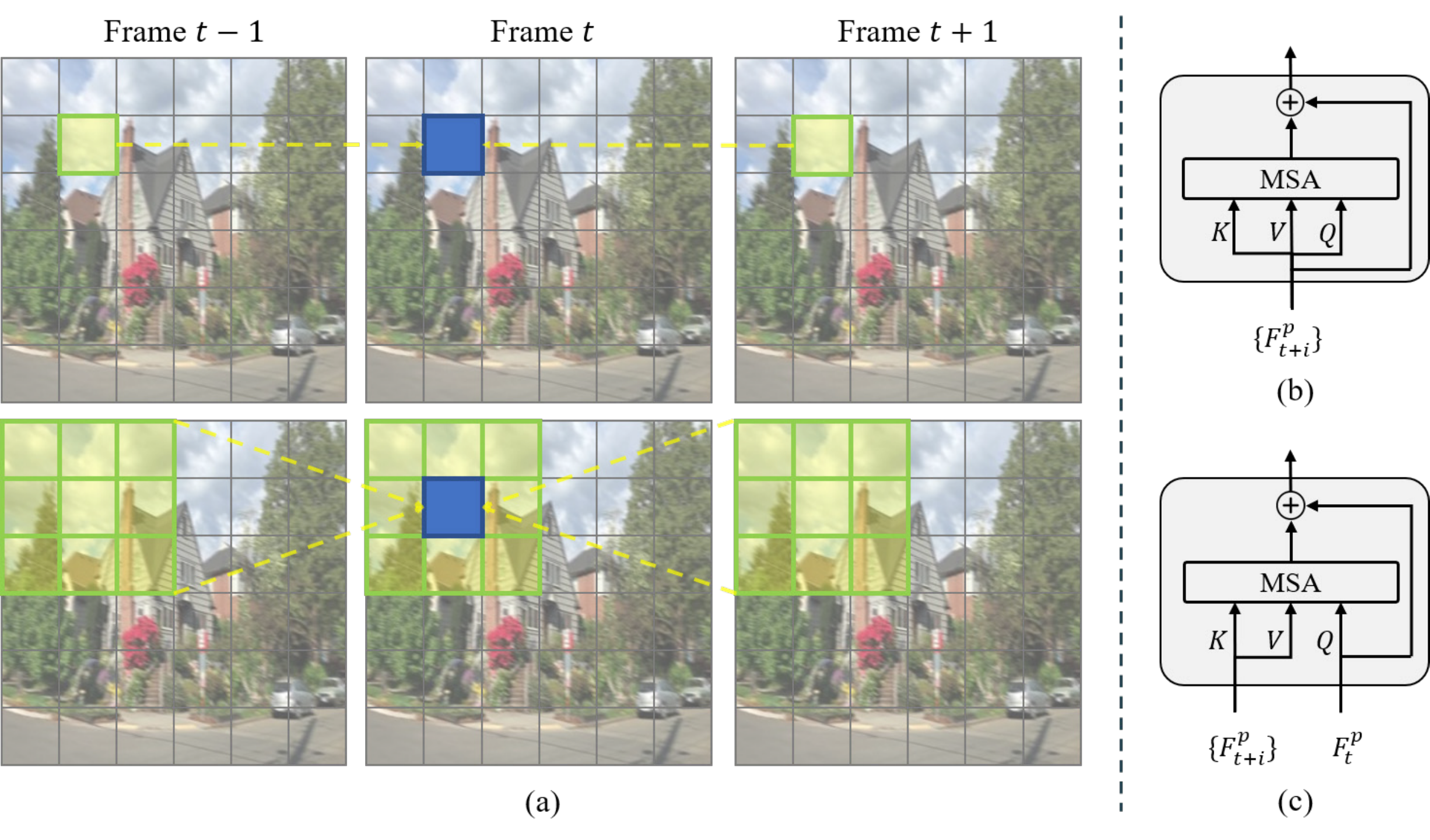}
    \caption{\textbf{(a)}: The visualization of the temporal attention schemes. The blue patch is the \textit{Query} and the yellow patches are the neighborhoods serving as \textit{Key} and \textit{Value} in self-attention. \textbf{(b):} Temporal attention within $p$-th spatial-temporal window. \textbf{(c):} Temporal cross-attention within $p$-th spatial-temporal window.}
    \label{fig:patch_temporal_transformer}
\end{figure}

\subsection{Temporal Modeling with Transformer}
Seeking complementary sharp information existing in adjacent frames can primarily improve deblurring performance. Unlike previous methods estimating the optical flows across multiple blurry frames, we exploit Transformer to build correspondence between reference and neighboring frames and aggregate these frames for the latent frame reconstruction by temporal attention. The basic approach would be performing temporal attention at the same position across multiple frames~(shown in the first row of Fig.~\ref{fig:patch_temporal_transformer}(a)) to gather sharp information from neighboring frames. However, we cannot obtain pleased restored results since the features of different frames are spatially misaligned. Without any aligning process, the misaligned features corresponding to blurry input frames may introduce some artifacts and deteriorate the deblurring performance. To tackle this problem, we try to build the correspondence with its spatial-temporal neighborhoods, performing temporal attention at patch level~(shown in the second row of Fig.~\ref{fig:patch_temporal_transformer}(a)) across frames. So the misaligned temporal features can be modeled and aggregated by the temporal Transformer effectively. 

Our temporal Transformer consists of temporal attention and temporal cross-attention modules. Specifically, the temporal attention module is illustrated in Fig.~\ref{fig:patch_temporal_transformer}(b). We first divide the multiple frame-level features $\{F_{t+i}\}^{N}_{i=-N}$ into spatio-temporal window feature $\{F_{t+i}^k\}$ with spatial size $m \times m$, and $k$ is the spatial window location index. So $k$-th window feature $\{F_{t+i}^k\}$ contains $m \times m \times (2N+1)$ tokens, and all tokens are unused to generate Query, Key and Values. Then we perform attention according to Eq.~\ref{eq:attn} with the generated embedding. The successive temporal cross-attention is shown in Fig.~\ref{fig:patch_temporal_transformer}{c}. Unlike temporal attention perform attention across all tokens, in temporal cross-attention, only the tokens in the reference feature map are used to generate Query, and the tokens in the spatio-temporal window are used to generate corresponding Keys and Values. Then the Q, K, and V are used to perform multi-head attention according to Eq.~\ref{eq:attn}. 

The temporal attention is performed to build correspondence mutually across all frames, enabling better information interaction across frames. At the same time, the temporal cross-attention can aggregate neighboring features to reference one effectively.

\subsection{Loss Function}
We employ the combination of Charbonnier loss $\mathcal{L}_{c}$ in~\cite{wang2019edvr} and the Perceptual loss $\mathcal{L}_{p}$ in~\cite{johnson2016perceptual, zhou2019spatio} as our optimization objective to obtain visual friendly restored results. In which,

\vspace{-0.5cm}
\begin{align}
    \mathcal{L}_{c} = \sqrt{{\| I_t - L_t \|}^2 + \epsilon},
\end{align}
 where $L_t$ is the restored latent frame, and $I_t$ is the corresponding ground truth frame. We empirically set $\epsilon = 0.001 $ for stable training. For the perceptual loss, which can generate more realist results, we adopt a pretrained VGG19~\cite{simonyanZ14vgg} to extract the intermediate features and adopt mean squared error~(MSE) to evaluate the distance of the features:
 
 \vspace{-0.5cm}
 \begin{align}
    \mathcal{L}_{p} = MSE(VGG_j(I_t), VGG_j(L_t)),
\end{align}
where $VGG_j$ means the output features of $j$-th layer. So the total loss are as follows:
\begin{align}
    \mathcal{L} = \mathcal{L}_{c} + \lambda \mathcal{L}_{p},
\end{align}
where $\lambda$ is default as $10^{-4}$ in our experiments.

\section{Experiments}

\subsection{Experimental Settings}
\paragraph{Datasets}
We conduct experiments on the popular video deblurring datasets DVD~\cite{su2017deep}, GOPRO~\cite{nah2017deep}, REDS~\cite{Nah_2019_CVPR_Workshops_REDS} and BSD~\cite{zhong2020estrnn}. The details of the data configurations are shown in Tab.~\ref{tab:dataset_description}.
DVD contains 71 blurry videos and the corresponding sharp videos. In which, 61 videos are used for training and 10 videos for validation. We follow the same configuration in~\cite{su2017deep}, which evaluates 10 frames in each test video. GOPRO contains 2,103 training frames from 22 sequences and 1,111 evaluation samples from 11 sequences. REDS is a large-scale dataset for both video super-resolution and deblurring. It contains 300 videos~(30000 frames) in total, and 240 for training, 30 for validation and the left 30 for testing. While the aforementioned datasets are all synthesized, we also evaluate VDTR on the recently proposed real-world video deblurring dataset BSD, where 60, 20, 20 videos are used for training, validation and testing, respectively.

\begin{table}[h]
    \centering
    \caption{Data configurations of DVD, GOPRO, REDS and BSD.}
    \label{tab:dataset_description}
    \resizebox{\linewidth}{!}{
    \begin{tabular}{|c|cc|cc|ccc|ccc|}
    \hline
    \multirow{2}{*}{Dataset} &\multicolumn{2}{c|}{GOPRO~\cite{nah2017deep}}  &\multicolumn{2}{c|}{DVD~\cite{su2017deep}} &  \multicolumn{3}{c|}{REDS~\cite{Nah_2019_CVPR_Workshops_REDS}} & \multicolumn{3}{c|}{BSD~\cite{zhong2020estrnn}}  \\ \cline{2-11}
            & Train   & Test  &  Train  & Test  & Train  &  Val  &  Test  & Train  & Val  & Test \\
    \hline \hline
    \multirow{2}{*}{Videos} 
        & 22   & 11   & 61   & 10   &  240  &  30  & 30  &  60  & 20  &  20  \\ \cline{2-11}
        & \multicolumn{2}{c|}{33}& \multicolumn{2}{c|}{71} & \multicolumn{3}{c|}{300}&\multicolumn{3}{c|}{100}  \\
    \hline
    \multirow{2}{*}{Frames} 
        & 2103 & 1111 & 5708 & 1000 &  24000& 3000 & 3000& 6000 & 2000& 3000 \\ \cline{2-11}
        & \multicolumn{2}{c|}{3214}&\multicolumn{2}{c|}{6708} & \multicolumn{3}{c|}{30000} & \multicolumn{3}{c|}{11000} \\
    \hline
    \end{tabular}
    }
\end{table}

\begin{table*}
    \centering
    \small
    \caption{The quantitative comparisons between the convolution-based methods and proposed VDTR on the synthesized video deblurring datasets DVD~\cite{su2017deep}, GOPRO~\cite{nah2017deep} and REDS~\cite{Nah_2019_CVPR_Workshops_REDS}. Our method achieves the highest PSNR and SSIM. $*$ denotes the results reported in~\cite{son2021recurrent}. The best is \textbf{highlighted} in bold, and the second best is \underline{underlined}.}
    \begin{tabular}{|c|p{2cm}<{\centering}p{2cm}<{\centering}|p{2cm}<{\centering}p{2cm}<{\centering}|p{2cm}<{\centering}p{2cm}<{\centering}|@{}}
    \hline
    \multirow{2}{*}{Methods} & \multicolumn{2}{c|}{DVD} & \multicolumn{2}{c|}{GOPRO} & \multicolumn{2}{c|}{REDS}      \\ \cline{2-7}
                           &  PSNR      &  SSIM      & PSNR       & SSIM    & PSNR       & SSIM    \\
    \hline\hline
    SRN~(CVPR2018)         & 30.53      & 0.894      & 30.61      & 0.908   & 31.21  &  0.8992  \\
    \hline
    DBN~(CVPR2017)         & 30.5       & 0.8844     & 29.31      & 0.8823  & 31.53  & 0.9030  \\
    DBLRNet~(TIP2018)      & 30.43      & 0.8862     & 28.24      & 0.8476  &  30.77  &  0.8890  \\
    IFIRNN~(CVPR2019)*    & 30.80      & 0.8991     & 29.97      & 0.8859  & 31.42  &  0.9041 \\
    STFAN~(ICCV2019)       & 31.05      & 0.9051     & 30.48      & 0.9026  & 31.59  & 0.9048 \\
    EDVR~(CVPRW2019)*      & 31.82      & 0.916      & 31.54      & 0.926   & \underline{34.80}  &  \underline{0.9487} \\
    ESTRNN~(ECCV2020)      & 30.68      & 0.8968     & 31.07      & 0.9023  & 31.93  & 0.9125 \\
    CDVDTSP~(CVPR2020)     & 32.13      & 0.9256     & \underline{31.67}      & \underline{0.9279}  & 32.03  &  0.9161 \\
    PVDNet~(TOG2021)*      & \underline{32.31}      & \underline{0.926}      & 31.52      & 0.921   & 32.11  & 0.9141  \\
    \hline
    VDTR (ours) & \textbf{33.13}& \textbf{0.9359} & \textbf{33.15}  & \textbf{0.9402}& \textbf{35.36} & \textbf{0.9525} \\
    \hline
    \end{tabular}
    \label{tab:results_on_dvd_gopro}
\end{table*}

\paragraph{Implementation Details}
During training, the input of VDTR is a short video clip consisting of 5 consecutive blurry frames (\ie,\ $N=2$) in RGB format. We perform random crop~(patch size is $256 \times 256$) and flip~(both horizontal and vertical) for the data augmentation. We employ the ADAM~\cite{Kingma2015adam} optimizer with $\beta_{1}=0.9$ and $\beta_{2}=0.99$. The initial learning-rate is $4\times10^{-4}$. All experiments are conducted on 8 NVIDIA Tesla V100 GPU with 32G memory.

\paragraph{Evaluation Metrics}
We compared VDTR with state-of-the-art convolution-based networks, including single image deblurring methods SRN~\cite{tao2018scale}, video deblurring methods DBN~\cite{su2017deep}, DBLRNet~\cite{zhang2018adversarial}, IFIRNN~\cite{nah2019recurrent}, EDVR~\cite{wang2019edvr}, STFAN~\cite{zhou2019spatio},  ESTRNN~\cite{zhong2020estrnn}, CDVDTSP~\cite{pan2020cascaded}, PVDNet~\cite{son2021recurrent} quantitatively and qualitatively. We adopt public available source codes for evaluation. Both Peak signal-to-noise ratio (PSNR) and structure similarity (SSIM) are adopted as evaluation metrics. 

\begin{figure*}[!htbp]
    \centering
    \begin{tabular}{@{}c@{}}
    \includegraphics[width=\linewidth]{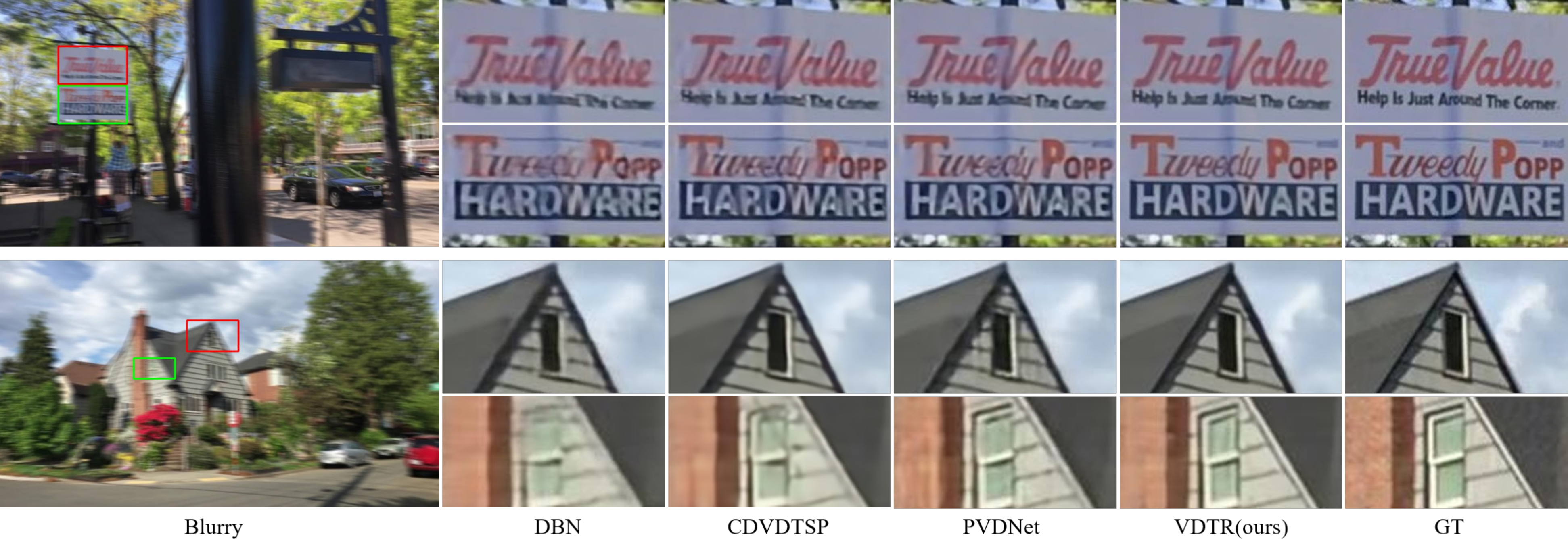}
    \end{tabular}
    \caption{The visual quality comparison to the classical CNN-based method DBN~\cite{su2017deep} and the state-of-the-art CNN-based video deblurring methods CDVDTSP~\cite{pan2020cascaded} and PVDNet~\cite{son2021recurrent}~(these two methods achieved highest PSNRs and SSIMs) on DVD~\cite{su2017deep} datasets. VDTR demonstrates strong competitiveness. For those highly blurry videos, the proposed VDTR can restore more details with the help of long-range modeling capacity and efficient temporal modeling with the Transformer. The frames are zoomed in for the best view. }
    \label{fig:quality_comparison_dvd}
\end{figure*}

\begin{figure*}[!htbp]
    \centering
    \begin{tabular}{@{}c@{}}
    \includegraphics[width=\linewidth]{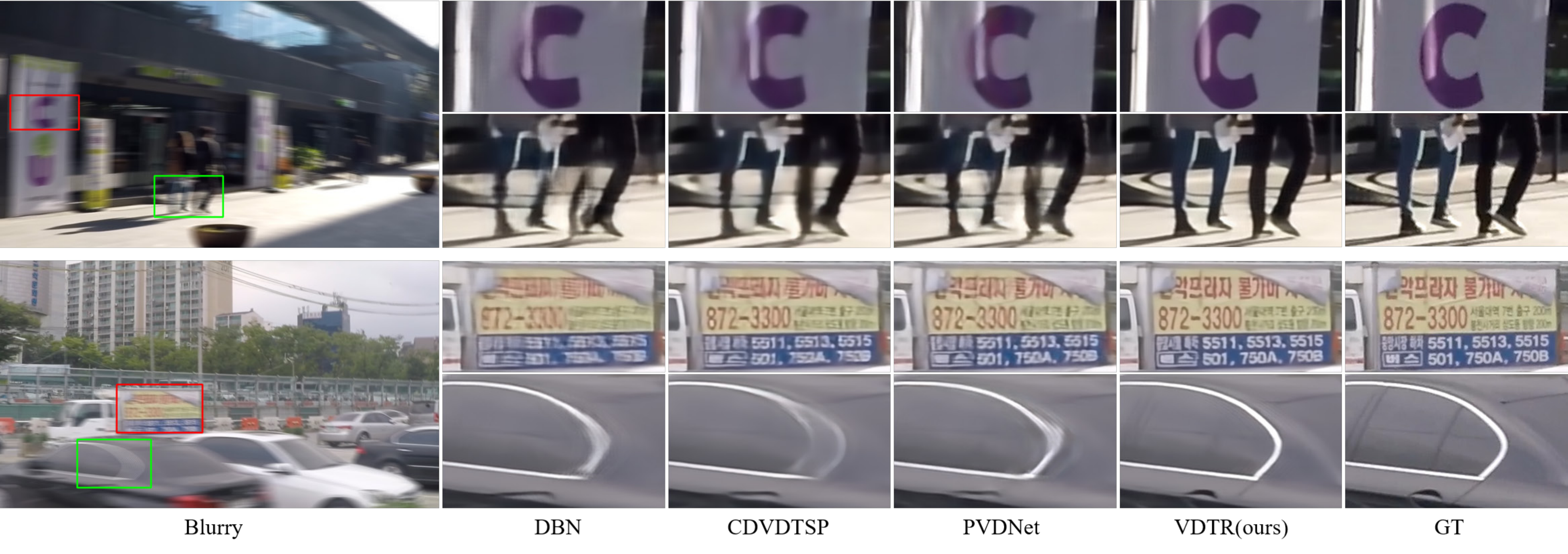}
    \end{tabular}
    \caption{The visual quality comparison to the classical CNN-based method DBN~\cite{su2017deep} and state-of-the-art CNN-based video deblurring methods CDVDTSP~\cite{pan2020cascaded} and PVDNet~\cite{son2021recurrent}~(these two methods achieved highest PSNRs and SSIMs) on GOPRO~\cite{nah2017deep} datasets. VDTR demonstrates strong competitiveness. The frames are zoomed in for the best view. }
    \label{fig:quality_comparison_gopro}
\end{figure*}

\begin{figure*}[!htbp]
    \centering
    \begin{tabular}{@{}c@{}}
    \includegraphics[width=\linewidth]{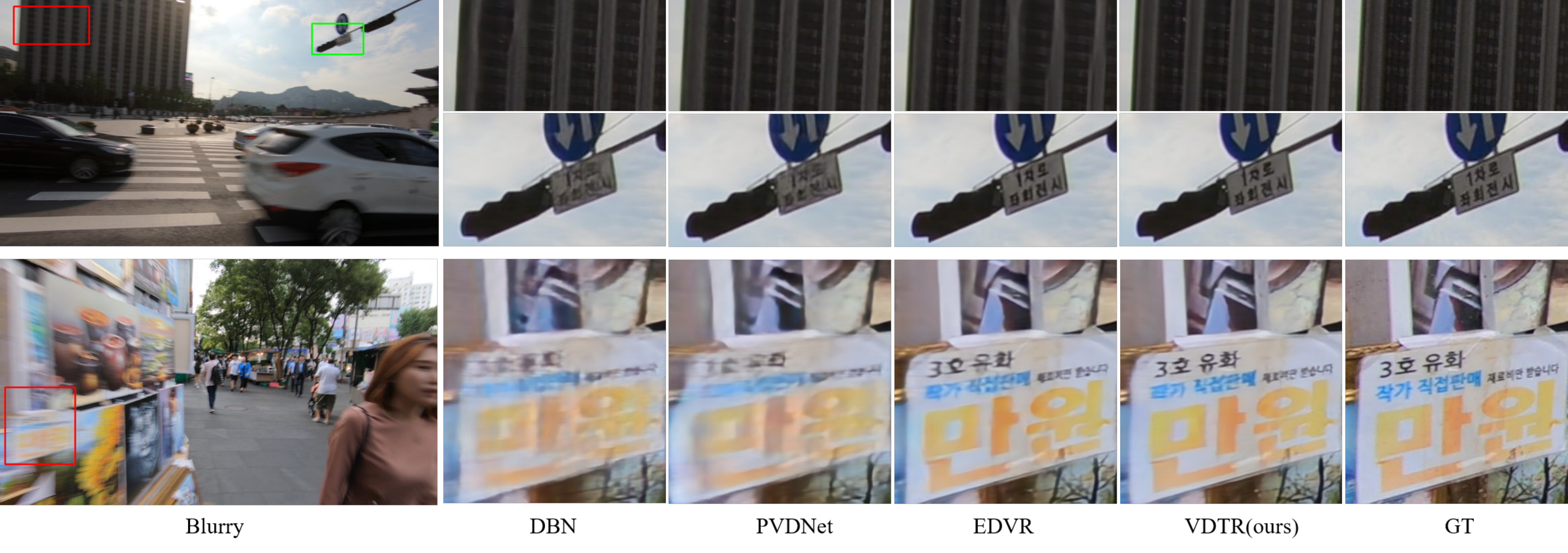}
    \end{tabular}
    \caption{The visual quality comparison to the classical CNN-based method DBN~\cite{su2017deep} and state-of-the-art CNN-based video deblurring methods PVDNet~\cite{son2021recurrent} and EDVR~\cite{wang2019edvr}~(these two methods achieved highest PSNRs and SSIMs) on REDS~\cite{Nah_2019_CVPR_Workshops_REDS} datasets. The frames are zoomed in for the best view. }
    \label{fig:quality_comparison_reds}
\end{figure*}

\begin{figure*}[h]
    \centering
    \begin{tabular}{@{}c@{}}
    \includegraphics[width=\linewidth]{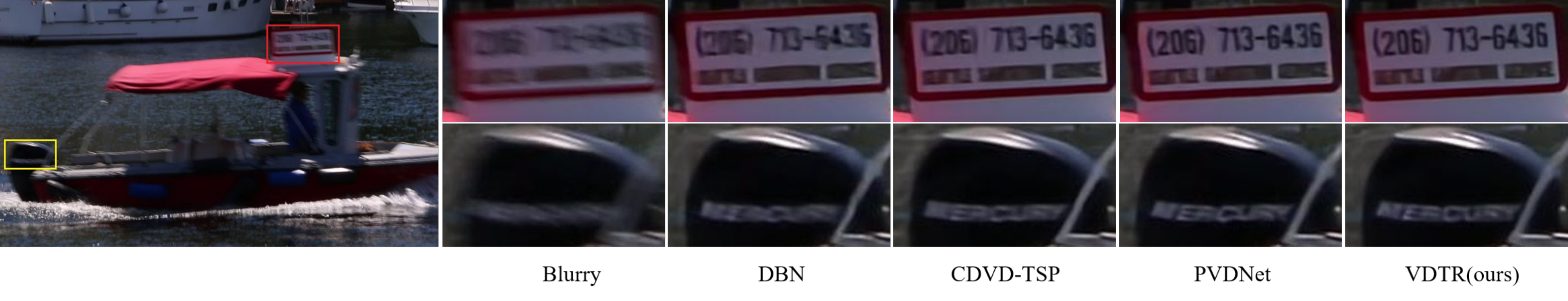}
    \end{tabular}
    \caption{The quality comparisons to state-of-the-art CNN-based video deblurring methods on the real blurry video~\cite{su2017deep}. The frames are zoomed in for the best view.}
    \label{fig:quality_comparison_real}
\end{figure*}

\begin{table*}[htbp]
    \centering
    \small
    \caption{The quantitative comparisons between proposed VDTR and CNN-based video deblurring methods on the real-world video delurring dataset BSD~\cite{zhong2020estrnn} with different data acquisition. The best is \textbf{highlighted}, and second best is \underline{underlined}.}
    \label{tab:results_on_bsd}
    \small
    \begin{tabular}{|c|p{2cm}<{\centering}p{2cm}<{\centering}|p{2cm}<{\centering}p{2cm}<{\centering}|p{2cm}<{\centering}p{2cm}<{\centering}|}
    \hline
    \multirow{2}{*}{Methods} & \multicolumn{2}{c|}{1ms-8ms} & \multicolumn{2}{c|}{2ms-16ms} & \multicolumn{2}{c|}{3ms-24ms} \\ \cline{2-7}
            &  PSNR  &  SSIM  & PSNR   & SSIM    & PSNR   & SSIM  \\
    \hline\hline
    SRN     &  31.84 & 0.917  & 29.95 & 0.891  & 28.92  &  0.882  \\
    \hline
    STRCNN  &  32.20 & 0.924  & 30.33 & 0.902  & 29.42  &  0.893  \\
    DBN     &  33.22 & 0.935  & 31.75 & 0.922  & 31.21  &  0.922  \\
    STFAN   &  32.78 & 0.9219 & 32.19 & 0.9187 & 29.47  &  0.8716 \\
    IFIRNN &  33.00 & 0.933  & 31.53 & 0.919  & 30.89  &  0.917   \\
    ESTRNN  &  {33.36} & 0.937  & 31.95 & 0.925  & 31.39  & 0.926  \\
    EDVR    &  33.16 & 0.9325 & 32.06 & 0.9158 & \underline{31.93}  & \underline{0.9261} \\
    CDVDTSP & \underline{33.54}  & \underline{0.9415} & 32.16  &  0.9261 & 31.58  & 0.9258 \\
    PVDNet  &  33.34 & {0.9371} & \underline{32.22} & \underline{0.9264} & 31.35  & 0.9226 \\
    \hline
    VDTR (ours) & \textbf{34.12} & \textbf{0.9436} & \textbf{32.69} & \textbf{0.9307}& \textbf{32.53} & \textbf{0.9363} \\
    \hline
    \end{tabular}
\end{table*}

\subsection{Results on Synthesized Dataset}
\paragraph{Quantitative Comparison}
Tab.~\ref{tab:results_on_dvd_gopro} shows the quantitative comparisons on the popular synthesized datasets. Thanks to the superior spatio-temporal information modeled by Transformer, VDTR achieves higher performance on these datasets than state-of-the-art CNN-based approaches in terms of both PSNR and SSIM. We note that PVDNet, CDVDTSP, and EDVR achieve state-of-the-art metrics on DVD, GOPRO, and REDS, respectively. However, the proposed VDTR model surpasses them with a significant margin. Taking the results on DVD into consideration, VDTR gets 0.7dB higher PSNR than the most competitive models, CDVD-TSP and PVDNet. Both two models adopt optical flow for temporal modeling. Compared to the IFIRNN, STFAN, and ESTRNN that adopt a recurrent architecture to restore the latent frame, VDTR reaches near 2.2dB, 1.9dB, and 2.4dB higher in terms of PSNR, respectively. These RNN-based models adopt fewer frames for temporal modeling, making the sharp information limited. VDTR also obtains 1.2dB higher PSNR compared to EDVR, which utilizes deformable convolution to align multiple frames. As for the quantitative results on GOPRO, VDTR provides the same excellent performance as on DVD, achieving 1.3dB higher PSNR than the previous state-of-the-art method CDVDTSP. Though EDVR obtaines much higher PSNR than other methods, our VDTR still surpasses it with 0.56dB PSNR. These superior quantitative comparisons significantly demonstrate the effectiveness of VDTR. 

\paragraph{Qualitative Comparison}
In addition to the superior performance on the evaluation metrics, we also provide some visual comparison results. Fig.~\ref{fig:quality_comparison_dvd}, Fig.~\ref{fig:quality_comparison_gopro} and Fig.~\ref{fig:quality_comparison_reds} shows the deblurred results in the test set of DVD, GOPRO, and REDS4~\cite{wang2019edvr}, and Fig.~\ref{fig:quality_comparison_real} shows the results on the real blurry videos provided by DVD. VDTR obtains high-quality results which look much clearer than other methods, for that long-range modeling for spatial modeling and relation modeling for temporal modeling are enabled by Transformer. In the extremely blurry scenarios, \eg, the car in the GOPRO shown in Fig.~\ref{fig:quality_comparison_gopro}, both CDVDTSP and PVDNet failed to restore the satisfying results though they obtained high average PSNRs and SSIMs in GOPRO dataset. We analyse the reasons for this as follows. CDVDTSP proposed a temporal sharpness prior to improving the deblurring performance, in which the optical flow estimation is not accurate when the input frames are blurry. Moreover, this temporal prior only focuses on a local blurry area, making it hard to handle the blurs caused by large motion. On the contrary, PVDNet introduces blur-invariant motion estimation and pixel volume-based motion compensation approaches to resolve the inaccurate optical flow estimation. Thus, it achieves better performance than CDVD-TSP, but some artifacts still exist in the restored results. We think such results come from the inefficient long-range dependencies modeling. Since VDTR adopts Transformer for both spatial and temporal modeling, the long-range dependencies and temporal variations can be extracted effectively. 

\begin{figure*}[!htbp]
    \centering
    \begin{tabular}{@{}c@{}}
    \includegraphics[width=\linewidth]{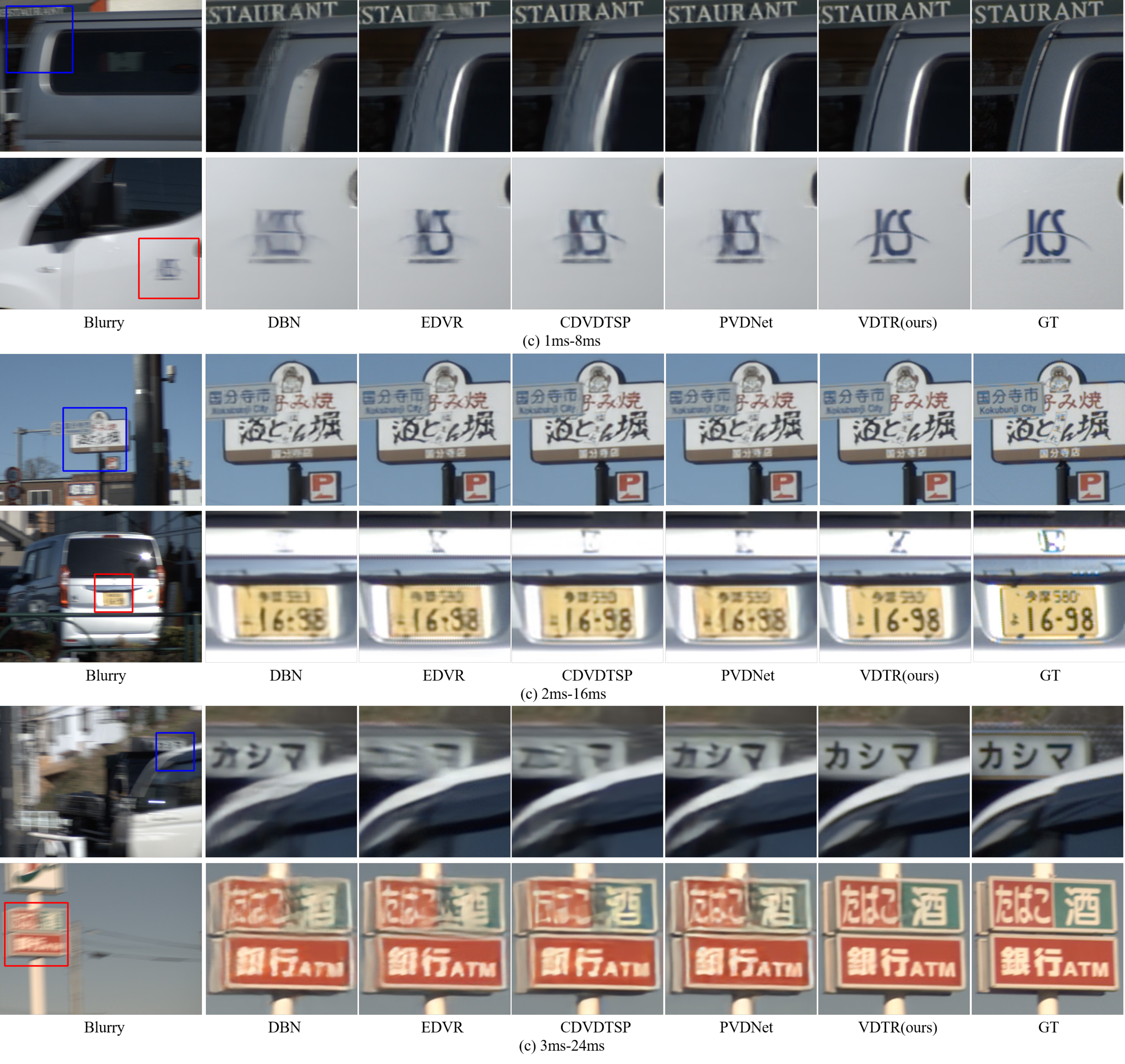}
    \end{tabular}
    \caption{The visual quality comparison to the classical CNN-based method DBN~\cite{su2017deep} and the state-of-the-art CNN-based video deblurring methods EDVR~\cite{wang2019edvr}, CDVDTSP~\cite{pan2020cascaded} and PVDNet~\cite{son2021recurrent}~(these methods achieved competitive PSNRs and SSIMs) on BSD~\cite{nah2017deep} datasets. VDTR demonstrates strong competitiveness. For those highly blurry videos, the proposed VDTR can restore more details with the help of long-range modeling capacity and efficient temporal modeling with the Transformer. The frames are zoomed in for the best view. }
    \label{fig:quality_comparison_bsd}
\end{figure*}

\subsection{Results on Real-world Dataset}
Compared to synthetic blurry video, the real-world blurry video contains more complicated blurs caused by various of motions, making real-world video deblurring more challenging. We evaluate VDTR on the recently proposed real-world video deblurring dataset BSD. Tab.~\ref{tab:results_on_bsd} shows the results of different data acquisition configurations, \ie, 1ms-8ms, 2ms-16ms, 3ms-24ms means different exposure times for capturing the paired sharp and blurry video. Usually, the videos in 3ms-24ms are more blurry with various motions and more challenging. VDTR achieves the best performance metrics in terms of PSNR and SSIM under different data acquisition configurations. Fig.~\ref{fig:quality_comparison_bsd} further shows some restored frames of VDTR and other CNN-based methods in a different subset. We see that VDTR performs favorably against previous CNN-based methods visually. These results are attributed to the Transformer-based multi-scale design for frame-level feature extraction and temporal Transformer for misaligned sharp information aggregation. VDTR can handle large blurs caused by various motions in the real blurry videos than CNN-based methods.

These quantitative and qualitative results on both synthesized and real-world datasets shown before reveal that VDTR surpasses the existing CNN-based video deblurring model and is worthy of further exploration.

\begin{table}[hbtp]
    \centering
    \small
    \caption{The effectiveness of Transformer-based encoder-decoder network and temporal Transformer. $G^{\#}_t$ and $G_f$ denote the temporal Transformer with different configurations and the Transformer-based encoder-decoder network, respectively. Specifically, $G^{1}_t$ contains only \textit{temporal attention} module with window size $4 \times 4$; $G^{2}_t$ and $G^{3}_t$ contain \textit{temporal attention and cross-attention} modules, while with $1 \times 1$ and $4\times 4$ spatial window size, respectively. The first row denote single frame deblurring. }
    \label{tab:ab_modules}
    \begin{tabular}{|c|cccc|cc|}
    \hline
    Model        & $G^{1}_t$ & $G^{2}_t$     & $G^{3}_t$ & $G_f$        & PSNR   & SSIM       \\ 
    \hline \hline
    Net1         &             &               &              &              & 30.01  & 0.8842   \\
    Net2         & $\checkmark$&               &              &              & 31.08  & 0.9026   \\
    Net3         &             &  $\checkmark$ &              &              & 31.23  & 0.9052   \\
    Net4         &             &               & $\checkmark$ &              & 31.42  & 0.9096   \\
    Net5         &             &               & $\checkmark$ & $\checkmark$ & 31.71  & 0.9139   \\
    \hline
    \end{tabular}
\end{table}

\subsection{Ablation Study}
To analyze the superior performance in quantitative and qualitative evaluation of VDTR, we conduct ablation studies on a more compact model, which follows the same architectural design and only contains a half number of channels and layers of VDTR. All the experiments in the following are taken on the DVD. 

\paragraph{Transformer-based Encoder-decoder Network}
The proposed Transformer-based encoder-decoder network can extract the spatial features with multi-scale information. We evaluate its effectiveness on Tab.~\ref{tab:ab_modules}, where net4 and net5 are without and with the proposed Transformer-based encoder-decoder network, respectively. We see that a near 0.3dB performance gain in terms of PSNR is obtained when adding the Encoder-decoder network for feature extraction.

\begin{figure}
    \centering
    \includegraphics[width=\linewidth]{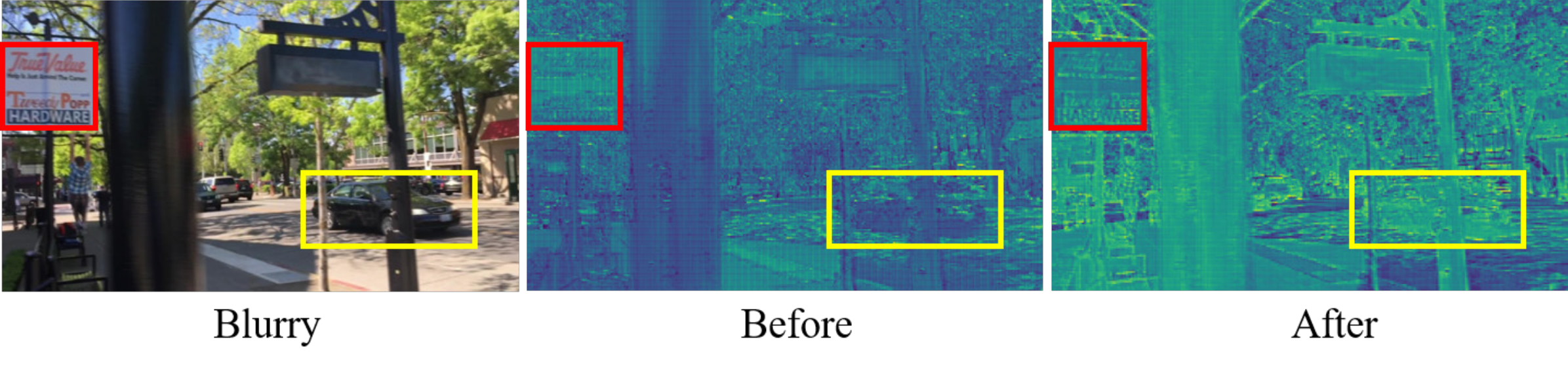}
    \caption{The visualizations of the intermediate features before and after temporal fusion $G_t$ in VDTR. \textbf{Left:} the reference blurry input frame. \textbf{Middle:} the features before temporal modeling~(the output of $G_f$). \textbf{Right:} the features after temporal modeling. Lighter regions indicate more significant changes. Obviously, the features are re-organized and focus on the most blurry areas after temporal modeling. }
    \label{fig:feats_comparison_vdtr_edvr}
\end{figure}

\paragraph{Temporal Transformer}
VDTR utilizes Transformer to build the correspondence across multiple temporal features implicitly. To validate the effectiveness of the temporal Transformer, we conduct different temporal attention-based experiments in Tab.~\ref{tab:ab_modules}, including single-frame deblurring without temporal modeling~(Net1), video deblurring with temporal Transformer $G_t$ of different configurations~(Net2, Net3, Net3). $G^1_t$ contains only the temporal cross-attention module with spatial window size $4 \times 4$ for temporal modeling. $G^2_t$ and $G^3_t$ consist of both temporal attention and cross attention with both window sizes $1\times 1$ and $4\times\ 4$, respectively. As for the ablation results, we see that the single frame deblurring strategy obtains the lowest PSNR because of the lack of complementary sharp information. The deblurring performance increases sharply when $G^1_t$ is added to model multiple temporal features. Furthermore, performing temporal attention and cross-attention can further successively boost the performance. The model with $G^3_t$ achieves the best performance among all strategies, because it can ease the misalignment problem across multiple frames by performing temporal aggregation in the spatio-temporal window rather than the same position across all frames.

We further visualize the immediate features of VDTR in Fig.~\ref{fig:feats_comparison_vdtr_edvr}. Intuitively, the features are re-organized and focus more on the most blurry regions. For example, the car in the yellow rectangle is dark before temporal modeling. It becomes lighter after temporal modeling, which means more changes will be added to this blurry region in the blurry frame. In contrast, the area in the red rectangle obtains fewer changes after temporal modeling because it is not as blurry as the car. Moreover, the edges in the feature map become sharper, which can be reconstructed as a clearer latent frame.

\begin{table}[btp]
    \caption{Comparison of local and global attention strategies. ``-P'' means the patch size during patch embedding process. ``-W'' stands for the window size of local attention. }
    \label{tab:ab_attention_methods}
    \centering
    \small
    \begin{tabular}{|c|cc|cc|}
    \hline
    Strategy  & Size-P & Size-W & PSNR & SSIM \\ 
    \hline \hline
    \multirow{2}{*}{Global} 
    & 16  &  N/A  & 29.51 & 0.8700  \\
    & 4   &  N/A  &  OOM  &  OOM    \\ 
    
    \hline
    \multirow{3}{*}{Local}
    & 16  &  4  & 30.03 & 0.8830  \\
    & 4   &  2  & 31.24 & 0.9063  \\
    & 4   &  4  & 31.42 & 0.9096  \\ 
    & 4   &  8  & 31.44 & 0.9101  \\
    \hline
    \end{tabular}
\end{table}

\begin{figure}[h]
    \centering
    \includegraphics[width=\linewidth]{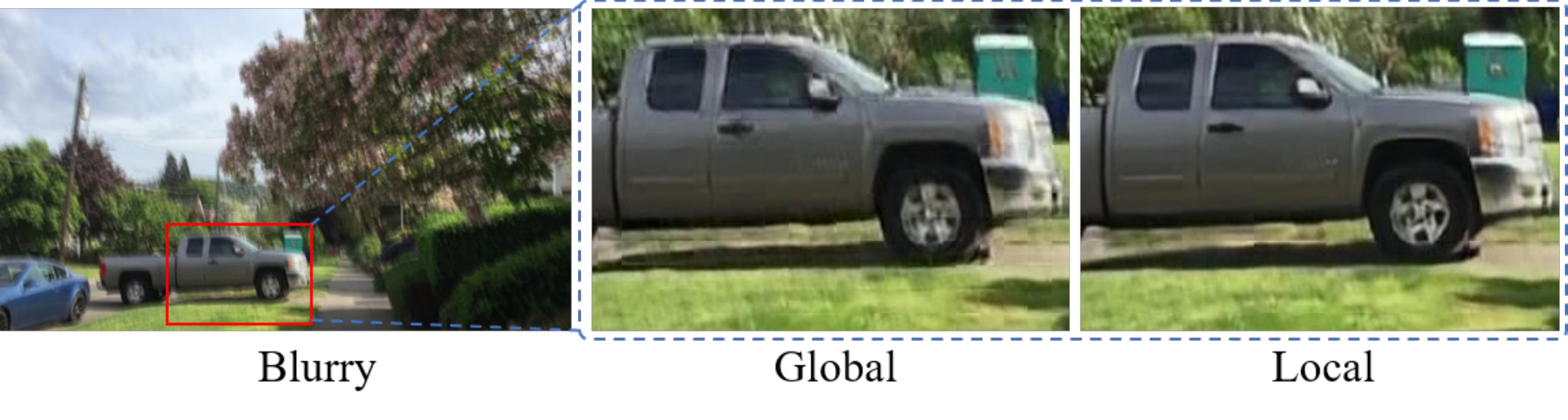}
    \caption{Visual quality comparison between global and local attention-based Transformers. }
    \label{fig:com_attn_strategy}
\end{figure}

\paragraph{Local Transformer}
To evaluate the effectiveness of the local attention mechanism, we conduct some experiments to explore local and global one. Tab.~\ref{tab:ab_attention_methods} shows the quantitative results with different configurations, and the visual quality comparisons are shown in the Fig~\ref{fig:com_attn_strategy}. As analyzed before, global attention cannot be directly used in the deblurring task because the high-resolution property of inputs results in unacceptable computational costs. So it is out of memory when a fine-grained frame embedding~(\ie, patch embedding with size $4 \times 4$) inputs the model. When the patch embedding size increases to 16, the global attention-based Transformer can work. However, it obtains the lowest metrics because it loses many details when the large patch partition severely downsamples the input frames during patch embedding. 

As for the local attention, high-resolution inputs can be handled. We perform patch embedding with patch size $4\times 4$ to generate high-resolution image embedding. The qualitative result of local Transformer in Tab.~\ref{tab:ab_attention_methods} shows a significant performance gain of 1.9dB in terms of PSNR than the global one. We think this performance gain comes from three parts: (1) The high-resolution inputs can maintain as much as high-frequency information for better restoration. (2) The local attention can better model the local-varying blurs in the blurry frames as CNNs have done, but the global attention is hard to model these local blur patterns. (3) It is easier to train a local attention-based model than a global one. We also analyze the influence of the size of non-overlapping windows shown in Tab.~\ref{tab:ab_attention_methods}. The performance will sharply rise when the window size increases from 2 to 4, then remain stable when it further increases to 8. So taking the computational cost and performance into consideration, we choose 4 as the window size. 

\begin{table}[h]
\centering
\small
\caption{The parameters and GFLOPs of the proposed VDTR and convolution-based methods. The GFLOPs are tested under the resolution $1280 \times 720$.}
\begin{tabular}{|c|cc|}
\hline
Methods   &  Param. (M)  &  GFLOPs  \\
\hline\hline
SRN       & 10.16    & 1441.9  \\
\hline
DBN       & 15.3     & 783.9   \\
DBLRNet   & 1.89     & 1740.0  \\
IFIRNN    & 12.2     & 1425    \\
STFAN     & 5.37     & 2833.1  \\
EDVR      & 23.6     & 2740.7  \\
ESTRNN    & 2.47     & 749.64  \\
CDVDTSP   & 15.3     & 4502.8   \\
PVDNet    & 10.5     & 1003.5  \\
\hline
VDTR~(ours) & 23.2     & 2244.7  \\
\hline
\end{tabular}
\label{tab:flops_params}
\end{table}

\paragraph{Model Efficiency}
We also compare the model efficiency~(number of parameters and GFLOPs) between VDTR and existing CNN-based video deblurring methods shown in Tab.~\ref{tab:flops_params}. VDTR achieves the best PSNR and SSIM on multiple popular video deblurring datasets while owning moderate parameters and GFLOPs compared to the state-of-the-art approaches. Specifically, VDTR has about the same number of parameters compared to EDVR but has 20\% fewer GFLOPs. Though ESTRNN and PVDNet possess fewer GFLOPs than VDTR, there is still a certain performance gap between them. Compared to CDVD-TSP, which obtained competitive video deblurring performance on the GOPRO dataset, VDTR surpasses more than 1.3dB with only 60\% less computational costs. 

\subsection{Discussions}
We also note that PVDNet, GOPRO and EDVR achieves state-of-the-art PSNR and SSIM on DVD, GOPRO, REDS, and BSD, shown in Tab.~\ref{tab:results_on_dvd_gopro} and Tab.~\ref{tab:results_on_bsd}. However, these CNN-based methods show a limited capacity to deal with different blurs well. We take the results in the real-world deblurring as an example. In Tab.~\ref{tab:results_on_bsd}, PVDNet obtains the second-best in 2ms-16ms, but it ranks 5th in 3ms-24ms, the same as EDVR. Compared to these CNN-based methods, the proposed VDTR achieves the best PSNR and SSIM on all acquisition conditions due to the powerful spatio-temporal modeling capacities of Transformer.

However, there still exist some open problems in video deblurring Transformer. 
For instance, we follow the patch embedding method and position encoding strategy designed for image classification in the original ViT, which may be sub-optimal for video deblurring. We believe it could be possible and more efficient to design task-specific embedding and position encoding for better performance.

\section{Conclusion}
In this paper, we propose to utilize Transformer for video deblurring. In contrast to previous CNN-based methods, the proposed VDTR utilizes superior long-range and relation modeling capacities for both spatial and temporal modeling, showing impressive improvements to learn spatio-temporal representations for video deblurring. We conduct extensive evaluations on both popular synthetic and real-world video deblurring benchmarks, and VDTR outperforms previous state-of-the-art with a significant performance gain in terms of both evaluation metrics and visual quality. Besides, the ablation study demonstrates the design of each module in VDTR. In the future, we plan to exploit novel Transformer architectures for more efficient video deblurring. We hope the proposed VDTR can be served as an alternative baseline model for video deblurring by utilizing the powerful modeling capabilities of Transformer for spatial and temporal modeling.

\section*{Acknowledgments}
We especially thank Jiahao Wang and Fei Yin for their discussions and feedback on the manuscript. This research was supported by the Key Program of the National Natural Science Foundation of China under Grant No. U1903213, the Shenzhen Key Laboratory of Marine IntelliSense and Computation under Contract ZDSYS20200811142605016.

\bibliography{ref}{}
\bibliographystyle{IEEEtran}

\vfill

\end{document}